\documentclass[sigconf]{acmart}

\usepackage{algorithm}
\usepackage{algorithmic}
\usepackage{multirow}
\usepackage{colortbl} 
\usepackage{balance}

\AtBeginDocument{%
  }

\copyrightyear{2026}
\acmYear{2026}
\setcopyright{cc}
\setcctype{by}
\acmConference[KDD '26]{Proceedings of the 32nd ACM SIGKDD Conference on Knowledge Discovery and Data Mining V.2}{August 09--13, 2026}{Jeju Island, Republic of Korea}
\acmBooktitle{Proceedings of the 32nd ACM SIGKDD Conference on Knowledge Discovery and Data Mining V.2 (KDD '26), August 09--13, 2026, Jeju Island, Republic of Korea}
\acmDOI{10.1145/3770855.3818848}
\acmISBN{979-8-4007-2259-2/2026/08}

\begin{document}

\title{MemNovo: Look Back at the Spectrum for Balanced \textit{De Novo} Peptide Sequencing from Mass Spectrometry}

\author{Dongxin Lyu}
\email{lyudongxin@westlake.edu.cn}
\orcid{0009-0004-4223-0557}
\affiliation{%
  \institution{Westlake University}
  \city{Hangzhou}
  \state{Zhejiang}
  \country{China}
}

\author{Jingbo Zhou}
\email{zhoujingbo@westlake.edu.cn}
\affiliation{%
  \institution{Westlake University}
  \city{Hangzhou}
  \state{Zhejiang}
  \country{China}
}
\author{Hongxin Xiang}
\email{xianghx@hnu.edu.cn}
\affiliation{%
  \institution{Hunan University}
  \city{Changsha}
  \state{Hunan}
  \country{China}
}
\author{Yuqiang Li}
\email{liyuqiang@pjlab.org.cn}
\affiliation{%
  \institution{Shanghai Artificial Intelligence Laboratory}
  \city{Shanghai}
  \country{China}
}

\author{Jun Xia}
\email{junxia@hkust-gz.edu.cn}
\affiliation{%
  \institution{HKUST-GZ \& HKUST}
  \city{Guangzhou}
  \state{Guangdong}
  \country{China}
}
\begin{abstract}
\textit{De novo} peptide sequencing from tandem mass spectrometry is pivotal in proteomics, enabling identification of novel peptides without reference databases. While recent Transformer-based encoder-decoder models have achieved remarkable performance, we uncover a critical pathology in their inference dynamics. Through comprehensive feature scaling experiments, we demonstrate that existing auto-regressive peptide decoders tend to over-rely on generated-sequence priors while progressively under-utilizing fine-grained physical evidence from the input mass spectrum. This phenomenon leads to suboptimal results, where generated peptide sequences are biologically plausible yet not faithful to the input spectrum. To rectify this, we propose \textbf{MemNovo}, a training-free and plug-and-play mechanism that re-balances peptide and spectral contributions at inference time. MemNovo alleviates the information bottleneck by establishing a persistent spectral memory bank and injecting retrieved features directly into the final decoding stage via an ultra-conservative residual connection. Theoretical analysis confirms that this mechanism restores the mutual information between the decoder state and the raw spectrum. Extensive experiments on the Nine Species benchmark with two representative baselines, Casanovo and InstaNovo, demonstrate that MemNovo consistently improves both amino acid precision and peptide precision, achieving up to 39.1\% relative improvement in peptide precision for Casanovo and up to 3.9\% for InstaNovo, with negligible computational overhead. Our code is available at \href{https://github.com/AIMS-Lab-HKUSTGZ/MemNovo}{here}.
\end{abstract}


\maketitle


\begin{figure}[ht]
    \centering
    \includegraphics[width=\linewidth]{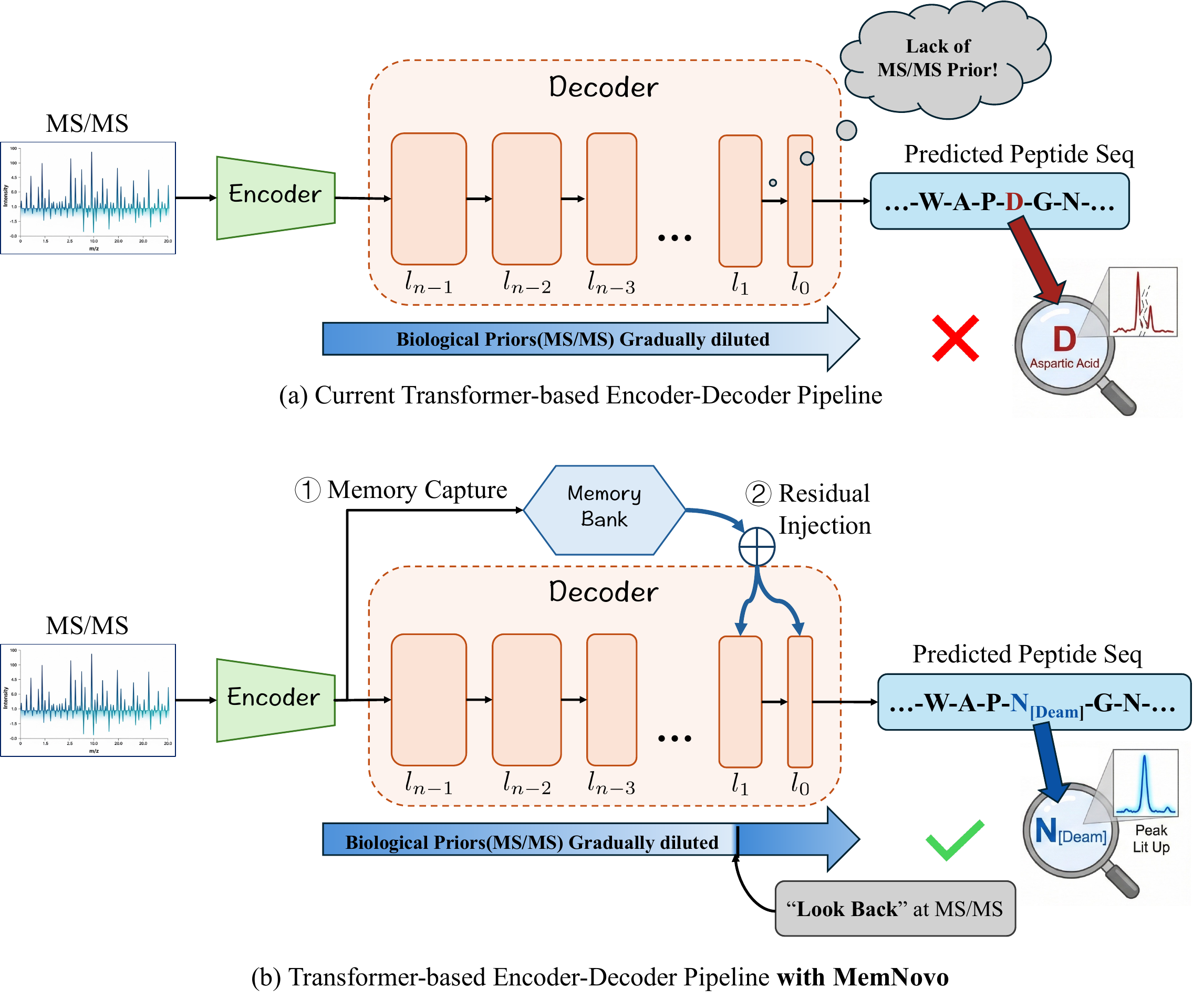}
    \caption{Spectral Under-utilization (top) and MemNovo's correction (bottom). }
    \Description{Two-panel comparison showing a mass spectrum with baseline decoder predicting an incorrect peptide sequence (top) and MemNovo correcting the prediction by re-accessing spectral peaks (bottom).}
    \label{fig:teaser}
\end{figure}

\section{Introduction}

Tandem mass spectrometry (MS/MS) has emerged as the premier high-throughput technology for protein identification and quantification. In bottom-up proteomics, peptides are digested, ionized, and fragmented into spectra that serve as fingerprints for sequence determination. While database search methods remain standard for identifying known proteins, they are inherently blind to novel variants, mutations, and organisms lacking reference genomes. \textit{De novo} peptide sequencing addresses this limitation by deriving amino acid sequences directly from raw spectral data, representing the \textit{holy grail} of computational proteomics.

The field has witnessed a paradigm shift with deep learning, particularly Transformer-based encoder-decoder architectures such as Casanovo~\cite{yilmaz2022novo} and InstaNovo~\cite{eloff2025instanovo}. These models treat peptide sequencing as a multimodal translation task, mapping mass spectral peaks and precursor information to amino acid sequences. Despite their state-of-the-art performance, we identify a fundamental vulnerability in their decision-making process. As autoregressive generation proceeds, the decoder relies increasingly on the history of generated amino acids to maintain biological coherence. We hypothesize that this creates an \textit{Information Bottleneck} where the model progressively ignores fine-grained mass spectral evidence encoded in the initial stages, a phenomenon we term \textit{Spectral Under-utilization}. As illustrated in Figure~\ref{fig:teaser}, \textit{de novo} sequencing models would predict a biologically plausible peptide sequence unsupported by the physical mass spectral signal, thus leading to suboptimal results.

To empirically validate this hypothesis, we conduct comprehensive feature scaling experiments that perturb feature magnitudes during inference to quantify model sensitivity to peptide and spectral inputs (see Figure~\ref{fig:sensitivity_imbalance} in \S\ref{sec:verification}). The results reveal a prevalent sensitivity imbalance in existing models: performance is highly sensitive to peptide-input perturbations but surprisingly robust to spectral-feature degradation. This confirms that the decision boundary is dominated by biological peptide priors rather than spectral evidence. This imbalance is particularly detrimental in scientific domains where fidelity to experimental evidence is paramount.

To address this issue, we propose \textbf{MemNovo}, a training-free and plug-and-play solution that can be integrated into any pre-trained Transformer decoder. Inspired by cognitive theories of memory re-access~\cite{zou2024look}, MemNovo allows the decoder to explicitly \textit{look back} at the original spectral input during amino acid generation. By caching the encoder output as a persistent memory bank and injecting retrieved features via an ultra-conservative residual connection, MemNovo forces the model to ground its predictions in the raw signal without disrupting learned representations.

Our contributions are summarized as follows.
\begin{itemize}
    \item \textbf{Discovery of Sensitivity Imbalance.} We propose the Sensitivity Scaling Framework and quantitatively demonstrate that current \textit{de novo} sequencing models suffer from severe peptide dominance, with sensitivity ratios up to 15$\times$, leading to spectral under-utilization and prediction errors.
    \item \textbf{Effective and Efficient MemNovo Framework.} We introduce a lightweight, inference-time mechanism that restores sensitivity balance. By enabling direct memory access to the spectral encoding via projection-free cross-attention, MemNovo enhances the mutual information between the generated hypothesis and the experimental data.
    \item \textbf{State-of-the-art Results.} We validate MemNovo on the standard Nine Species benchmark using both Casanovo and InstaNovo. The results show consistent improvements in amino acid precision and peptide precision, with up to +39.1\% relative improvement in peptide precision for Casanovo and only $\sim$1\% increase in computational latency.
\end{itemize}

\section{Related Work}

\subsection{Database Search Methods for Peptide Identification}

Database search has long been the dominant paradigm for peptide identification from tandem mass spectra. In this approach, experimental spectra are compared against theoretical spectra generated from in silico digestion of a protein sequence database, and statistical scoring functions are used to rank candidate peptide-spectrum matches (PSMs).

SEQUEST~\cite{eng1994approach}, one of the earliest database search engines developed in 1994, pioneered the cross-correlation scoring approach by comparing experimental and theoretical spectra. Mascot~\cite{perkins1999probability} subsequently introduced a probabilistic scoring model based on the MOWSE algorithm, becoming one of the most widely used commercial search engines. X!Tandem~\cite{craig2004tandem} offered an open-source alternative with support for iterative refinement searches. The MaxQuant platform~\cite{cox2008maxquant}, powered by its integrated Andromeda~\cite{cox2011andromeda} search engine, became a standard tool for label-free quantitative proteomics by combining probabilistic peptide scoring with sophisticated quantification algorithms. More recently, MSFragger~\cite{kong2017msfragger} introduced a fragment-ion indexing strategy that achieved over 100-fold speedup compared to conventional tools, enabling practical open database searches for comprehensive identification of post-translational modifications. Other notable engines include Comet~\cite{eng2013comet}, MS-GF+~\cite{kim2014ms}, and PEAKS~\cite{zhang2012peaks}, each contributing distinct scoring strategies and search capabilities.

While these database search methods have been continuously refined over the past three decades, they share a fundamental limitation, namely their reliance on pre-existing protein sequence databases, which restricts their ability to identify novel peptides, splice variants, or proteins from organisms with incomplete genome annotations. This inherent constraint motivates the development of \textit{de novo} peptide sequencing approaches that derive amino acid sequences directly from spectral data.

\subsection{Deep Learning-Based \textit{De Novo} Peptide Sequencing}

The application of deep learning to \textit{de novo} peptide sequencing has fundamentally reshaped the field~\cite{xia2025comprehensive}. DeepNovo~\cite{tran2017novo} was the pioneering work that introduced a CNN-LSTM hybrid architecture for predicting peptide sequences directly from tandem mass spectra, achieving significant improvements over traditional methods such as PepNovo+~\cite{frank2005pepnovo}, Novor~\cite{ma2015novor}, and PEAKS~\cite{zhang2012peaks}. DeepNovo-DIA~\cite{tran2019deep} extended this framework to data-independent acquisition data. SMSNet~\cite{karunratanakul2019uncovering} adopted a similar CNN-RNN architecture with an excitation mechanism for identifying novel peptides.

PointNovo~\cite{qiao2021computationally} innovatively treated mass spectra as point clouds using an order-invariant neural network, achieving instrument-resolution independence. GraphNovo~\cite{mao2023mitigating} employed a two-stage graph neural network approach to address the missing-fragmentation problem, using a Graphormer encoder to find optimal paths in spectrum graphs.

The field converged on Transformer architectures with the introduction of Casanovo~\cite{yilmaz2022novo}, which pioneered the application of a Transformer encoder-decoder to translate raw mass spectra into peptide sequences without discretization of the $m/z$ axis. InstaNovo~\cite{eloff2025instanovo} enhanced this paradigm with multi-scale sinusoidal embeddings and a knapsack-constrained beam search for improved decoding.
Subsequent Transformer-based models have diversified along several axes. DPST~\cite{yang2022dpst} introduced amino-acid-aware attention with confidence value aggregation. $\pi$-HelixNovo~\cite{yang2024introducing} processes complementary synthetic spectra alongside experimental data through dual encoders. ContraNovo~\cite{jin2024contranovo} adopted contrastive learning to model pairwise spectra-peptide interactions and integrated prefix/suffix mass data during decoding. AdaNovo~\cite{xia2024adanovo} proposed conditional mutual information-based re-weighting to improve identification of amino acids with post-translational modifications. NovoB~\cite{lee2024bidirectional} introduced bidirectional decoding via twin decoders. Spectralis~\cite{klaproth2024deep} proposed fragment ion series classification for precise \textit{de novo} sequencing. PepNet~\cite{liu2022pepnet} presented a fully convolutional architecture for non-autoregressive prediction.

More recently, $\pi$-PrimeNovo~\cite{zhang2025pi} employed non-autoregressive decoding with parallel amino acid prediction and mass constraint verification, achieving substantial speedup. SearchNovo~\cite{xia2024bridging} and ReNovo~\cite{chen2025renovo} bridged database search and \textit{de novo} sequencing by leveraging retrieval-augmented strategies. RankNovo~\cite{qiu2025universal} and CrossNovo~\cite{zhang2025bidirectional} introduced universal reranking and knowledge distillation approaches, respectively.

Despite these architectural innovations, a critical gap persists in understanding how these models integrate information from different input sources. Prior research has primarily focused on optimizing training strategies and decoding speeds, leaving the internal dynamics of information balance, specifically whether models over-rely on certain inputs at the expense of others, largely unexplored. Our work addresses this fundamental limitation by systematically investigating input source contributions, moving beyond aggregate performance metrics to analyze the granular dependence on different information sources.

\section{Methods}

\subsection{Problem Formulation}

\noindent\textbf{Notation.}
Let $\mathcal{S} = \{(m_i, I_i)\}_{i=1}^{N}$ denote a tandem mass spectrum consisting of $N$ peaks, where $m_i$ and $I_i$ represent the mass-to-charge ratio ($m/z$) and intensity, respectively. Let $\mathcal{P} = (M, z)$ denote the precursor information, comprising the precursor mass $M$ and charge state $z$. The goal of \textit{de novo} sequencing is to map the tuple $(\mathcal{S}, \mathcal{P})$ to a peptide sequence $\mathbf{y} = (y_1, y_2, \ldots, y_L)$, where $y_t \in \mathcal{A}$ represents an amino acid residue from the standard alphabet $\mathcal{A}$.

\noindent\textbf{Baseline Architecture and The Sensitivity Imbalance Hypothesis.}
We select two representative models for validation, namely Casanovo~\cite{yilmaz2022novo}, a pioneering Transformer-based model for \textit{de novo} peptide sequencing, and InstaNovo~\cite{eloff2025instanovo}, a current state-of-the-art model. Both employ a Transformer encoder-decoder architecture. The encoder processes spectral peaks and precursor information to produce a latent representation $\mathbf{S}_{\text{enc}} = \text{Encoder}(\mathcal{S}, \mathcal{P}) \in \mathbb{R}^{B \times N \times d}$, where $B$ is the batch size and $d$ is the hidden dimension. The decoder generates the peptide sequence autoregressively, with the hidden state $\mathbf{H}^l$ at layer $l$ updated based on the previous tokens and the encoder output via standard cross-attention.

We hypothesize that this architecture suffers from \textit{Sensitivity Imbalance}, meaning that as the sequence length grows, the autoregressive decoder over-relies on the linguistic patterns of the generated peptide sequence (the \textit{peptide input}) while progressively neglecting the physical evidence from the spectrum (the \textit{spectrum input}). This \textit{Spectral Under-utilization} leads to predictions where the generated sequence is grammatically plausible but factually inconsistent with the raw spectral data.

\subsection{Diagnosing Sensitivity Imbalance via Inference-Time Scaling}

To validate the sensitivity imbalance hypothesis, we propose a \textbf{Sensitivity Scaling Framework}, an inference-time diagnostic tool inspired by recent work on multimodal analysis~\cite{zou2024look}. Unlike binary ablation studies that mask inputs entirely, our framework applies continuous scaling factors to feature magnitudes during inference to quantify the graded sensitivity of the model to each input source.

We define a scaling operator $\Phi(\mathbf{x}, \alpha) = \alpha \cdot \mathbf{x}$, and intervene in the latent space by scaling the encoder outputs for the spectrum ($\mathbf{S}_{\text{enc}}$) and the peptide-related features (including precursor and history embeddings, denoted as $\mathcal{P}_{\text{enc}}$) independently as follows.
\begin{equation}
    \mathbf{S}'_{\text{enc}} = \Phi(\mathbf{S}_{\text{enc}}, \alpha_s), \quad \mathcal{P}'_{\text{enc}} = \Phi(\mathcal{P}_{\text{enc}}, \alpha_p)
\end{equation}
Here $\alpha_s, \alpha_p$ are drawn from a set of scaling factors. We then measure the \textit{Sensitivity} ($\Delta$) as
\begin{equation}
\Delta_m = \mathbb{E}_{\alpha \neq 1}\left[\frac{|\text{Perf}(\alpha_m) - \text{Perf}(1.0)|}{\text{Perf}(1.0)}\right], \quad m \in \{s, p\}
\end{equation}
A high $\Delta_p$ combined with a low $\Delta_s$ indicates heavy reliance on peptide priors and failure to fully utilize spectral information, confirming the presence of sensitivity imbalance.

\noindent\textbf{Architecture-Dependent Scaling Granularity.}
Ideally, the scaling range should be consistent across models. However, we identify that the permissible perturbation range is strictly constrained by the normalization architecture. Casanovo employs a \textit{post-norm} architecture, which is structurally robust to magnitude shifts, allowing a wide logarithmic range ($\alpha \in [0.1, 10.0]$). In contrast, InstaNovo utilizes a \textit{pre-norm} design where feature magnitudes are directly coupled with the residual stream variance. We empirically observe that scaling factors beyond $\pm 1\%$ in InstaNovo lead to catastrophic manifold collapse, producing empty or repetitive sequences. Consequently, we restrict InstaNovo to a linear range of $\alpha \in [0.99, 1.01]$.

\noindent\textbf{Comparability of the Sensitivity Ratio.}
Although the absolute sensitivity values $\Delta_s$ and $\Delta_p$ are not directly comparable across models due to the different scaling ranges, the \textit{Sensitivity Ratio} $\Delta_p / \Delta_s$ remains a meaningful cross-model diagnostic. This is because the ratio cancels out the effect of the perturbation magnitude under the assumption that both modalities respond approximately linearly to small perturbations within their respective ranges. Concretely, if $\Delta_m \approx c_m \cdot |\delta|$ for each modality $m$ (where $\delta$ characterizes the perturbation strength), the ratio $\Delta_p / \Delta_s = c_p / c_s$ becomes range-invariant and reflects the intrinsic sensitivity imbalance of the model. We therefore report the Sensitivity Ratio as the primary cross-architecture metric.

\begin{figure*}
    \centering
    \includegraphics[width=\linewidth]{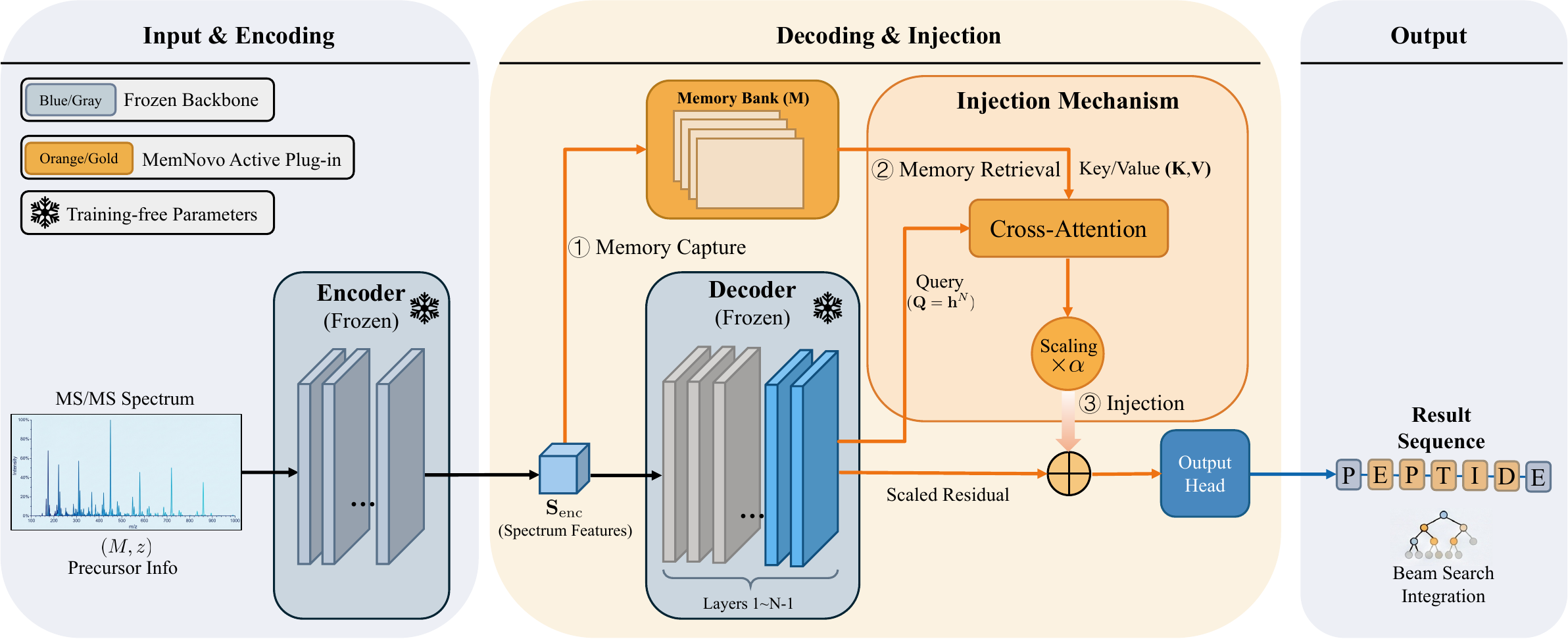}
    \caption{Overview of the MemNovo framework. The system comprises three stages. (Left) The tandem mass spectrum is processed by a frozen encoder to yield latent features $\mathbf{S}_{\text{enc}}$. (Middle) These features are cached in a persistent Memory Bank ($\mathbf{M}$). During the final layer of the frozen decoder, the Injection Mechanism performs projection-free memory retrieval via cross-attention and injects the spectral signal through an ultra-conservative residual connection (scaled by $\alpha$). (Right) The enhanced representation guides the Beam Search to produce the final peptide sequence, correcting potential prediction errors.}
    \Description{Three-stage pipeline diagram: a frozen encoder processes a mass spectrum into latent features, which are cached in a memory bank; the frozen decoder performs projection-free cross-attention retrieval from the memory bank with ultra-conservative residual injection; beam search produces the final peptide sequence.}
    \label{fig:framework}
\end{figure*}

\subsection{MemNovo: Memory-Enhanced Decoding}

To rectify the identified imbalance, we introduce \textbf{MemNovo}, a plug-and-play, training-free mechanism. MemNovo breaks the encoder bottleneck by allowing the decoder to \textit{retrace} the original spectral input directly during generation (Figure~\ref{fig:framework}). It consists of three key components.

\noindent\textbf{Persistent Memory Capture.}
Standard decoders view the encoder output as a static context. MemNovo explicitly treats it as a queryable memory bank. At the initialization of decoding, we capture the full spectral encoding and store it as a persistent memory $\mathbf{M} \leftarrow \mathbf{S}_{\text{enc}} \in \mathbb{R}^{B \times N \times d}$. This memory remains immutable throughout the decoding steps, ensuring that the original, fine-grained feature landscape is always accessible regardless of the decoding depth.

\noindent\textbf{Memory Retrieval.}
We inject a lightweight retrieval mechanism into the final $k$ layers of the decoder (empirically $k=1$ yields the best trade-off; see \S\ref{sec:ablation}). For the target layer $l$, given the current decoder hidden state $\mathbf{H}^l \in \mathbb{R}^{B \times L \times d}$, we compute a scaled dot-product similarity against the spectral memory as
\begin{equation}
    \mathbf{A}^l = \text{softmax}\!\left(\frac{\mathbf{H}^l (\mathbf{M})^\top}{\sqrt{d}}\right) \in \mathbb{R}^{B \times L \times N}
\end{equation}
The retrieved spectral context is then computed as $\mathbf{S}_{\text{pool}}^l = \mathbf{A}^l \cdot \mathbf{M}$.

\noindent\textbf{Distinction from Standard Cross-Attention.}
While each decoder layer already performs cross-attention over $\mathbf{S}_{\text{enc}}$, MemNovo's retrieval differs in three critical aspects. (i) \textit{No learned projections}. Standard cross-attention uses learned $W_Q, W_K, W_V$ matrices that may have been optimized to favor peptide-language features, inadvertently compressing spectral information. MemNovo computes similarity directly in the hidden space, bypassing this bottleneck. (ii) \textit{No post-processing}. The standard cross-attention output undergoes LayerNorm, FFN, and residual connections, each diluting the spectral signal (consistent with the Data Processing Inequality analysis in \S\ref{sec:theory}). MemNovo injects \textit{after} the final layer's processing, serving as a last-mile correction. (iii) \textit{Softmax normalization}. We adopt scaled softmax rather than the ReLU activation used in prior memory-retracing work~\cite{zou2024look}, as the normalized probability distribution prevents magnitude drift in the training-free setting and yields superior empirical performance.

\noindent\textbf{Ultra-Conservative Residual Injection.}
A critical challenge in training-free enhancement is stability. Direct intervention in deep Transformer layers can disrupt the learned representations. To mitigate this, MemNovo employs an ultra-conservative residual connection with a tiny scaling coefficient $\alpha$ as
\begin{equation}
    \mathbf{H}^l_{\text{enhanced}} = \mathbf{H}^l + \alpha \cdot \mathbf{S}_{\text{pool}}^l
\end{equation}
where $\alpha$ is a hyperparameter (default $\alpha=0.005$). This design ensures that the retrieved memory acts as a subtle nudge rather than a dominant signal, preserving the generative grammar learned by the base model while injecting precise spectral evidence to correct potential prediction errors.

\noindent\textbf{Implementation Details.}
MemNovo is designed for minimal overhead.
For \textit{beam search compatibility}, we handle dynamic batch expansion by broadcasting the memory bank $\mathbf{M}$ via \texttt{repeat\_interleave} to match the beam width $K$, ensuring alignment between hypotheses and source spectra without redundant re-encoding. The memory bank is expanded once per spectrum and reused across all beam search steps.
Regarding \textit{complexity}, the additional cross-attention operation has a time complexity of $O(L \cdot N \cdot d)$ per layer. Since we apply this only to the final layer ($k=1$), and the peptide length $L$ and number of peaks $N$ are typically small ($L \leq 30$, $N \leq 200$), the asymptotic complexity of the entire decoding process remains unchanged. The complete inference procedure is summarized in Algorithm~\ref{alg:memnovo}.

\begin{algorithm}[t]
\caption{MemNovo Inference Procedure}
\label{alg:memnovo}
\begin{algorithmic}[1]
\REQUIRE Pretrained encoder-decoder $f$, Spectrum $\mathcal{S}$, Precursor $\mathcal{P}$, Injection scale $\alpha=0.005$, Target layer $l^*=D$
\ENSURE Predicted peptide sequence $\mathbf{y}$
\STATE $\mathbf{S}_{\text{enc}} \gets f.\text{Encoder}(\mathcal{S}, \mathcal{P})$
\STATE $\mathbf{M} \gets \mathbf{S}_{\text{enc}}$
\STATE $\mathbf{y} \gets [\text{BOS}]$
\FOR{$t = 1, 2, \ldots, L_{\max}$}
    \STATE $\mathbf{H}^0 \gets \text{Embed}(\mathbf{y}_{<t})$
    \FOR{$l = 1, \ldots, D$}
        \STATE $\mathbf{H}^l \gets \text{Layer}_l(\mathbf{H}^{l-1}, \mathbf{S}_{\text{enc}})$
        \IF{$l = l^*$}
            \STATE $\mathbf{A}^l \gets \text{softmax}(\frac{\mathbf{H}^l \mathbf{M}^\top}{\sqrt{d}})$
            \STATE $\mathbf{S}_{\text{pool}}^l \gets \mathbf{A}^l \cdot \mathbf{M}$
            \STATE $\mathbf{H}^l \gets \mathbf{H}^l + \alpha \cdot \mathbf{S}_{\text{pool}}^l$
        \ENDIF
    \ENDFOR
    \STATE $y_t \gets \text{BeamSearch}(\text{Head}(\mathbf{H}^D))$
    \IF{$y_t = \text{EOS}$}
        \STATE \textbf{break}
    \ENDIF
\ENDFOR
\STATE \RETURN $\mathbf{y}$
\end{algorithmic}
\end{algorithm}

\subsection{Theoretical Analysis}
\label{sec:theory}

We analyze MemNovo's efficacy through the lens of information theory, demonstrating that our mechanism mitigates the information loss inherent in autoregressive decoding.

\noindent\textbf{Information Fading in Autoregressive Decoders.}
Let $\mathbf{H}^l$ denote the hidden state at layer $l$. By the Data Processing Inequality (DPI), successive processing layers cannot increase the mutual information with the original spectral input $\mathcal{S}$, that is,
\begin{equation}
    I(\mathbf{H}^l; \mathcal{S}) \leq I(\mathbf{H}^{l-1}; \mathcal{S})
\end{equation}
As the network depth increases and the sequence grows, the dependency on the initial spectral encoding naturally diminishes (\textit{Spectral Fading}), increasing the risk of prediction errors driven by the language prior $P(\mathbf{y}_{<t})$.

\noindent\textbf{Proposition 1 (Mutual Information Restoration).}
MemNovo introduces a direct retrieval path from the persistent memory $\mathbf{M}$ (where $\mathbf{M} \leftarrow \mathbf{S}_{\text{enc}}$) to the target layer. Let $\mathbf{H}^l_{\text{enh}}$ denote the enhanced hidden state. Since $\mathbf{H}^l_{\text{enh}} = f(\mathbf{H}^l, \mathbf{M})$ and $\mathbf{M}$ is a sufficient statistic of $\mathcal{S}$ at the encoder output, we have
\begin{equation}
    I(\mathbf{H}^l_{\text{enh}}; \mathcal{S}) \geq I(\mathbf{H}^l; \mathcal{S})
\end{equation}
\textit{Proof.} The injection operation re-introduces $\mathbf{M}$ into the computation graph. Since $I(\mathbf{H}^l_{\text{enh}}; \mathcal{S} | \mathbf{H}^l) \geq 0$ by non-negativity of conditional mutual information, the total spectral information content is preserved or increased. \hfill$\square$

\noindent\textbf{Proposition 2 (Conditional Entropy Reduction).}
Under the assumption that the ground-truth peptide $\mathbf{y}^*$ is a deterministic function of the spectrum $\mathcal{S}$, i.e., $H(\mathbf{y}^* | \mathcal{S}) = 0$, enhancing the spectral information in $\mathbf{H}^l$ reduces the predictive uncertainty as
\begin{equation}
    H(\mathbf{y}^* | \mathbf{H}^l_{\text{enh}}) \leq H(\mathbf{y}^* | \mathbf{H}^l)
\end{equation}
\textit{Proof.} Since $H(\mathbf{y}^* | \mathcal{S}) = 0$, we have $I(\mathbf{y}^*; \mathcal{S}) = H(\mathbf{y}^*)$, meaning $\mathbf{y}^*$ is fully determined by $\mathcal{S}$. This implies the Markov chain $\mathbf{y}^* \leftrightarrow \mathcal{S} \leftrightarrow \mathbf{H}^l$, and consequently $I(\mathbf{y}^*; \mathbf{H}^l) \leq I(\mathcal{S}; \mathbf{H}^l)$. From Proposition~1, $I(\mathcal{S}; \mathbf{H}^l_{\text{enh}}) \geq I(\mathcal{S}; \mathbf{H}^l)$, and by the Markov property $I(\mathbf{y}^*; \mathbf{H}^l_{\text{enh}}) \geq I(\mathbf{y}^*; \mathbf{H}^l)$. Since $H(\mathbf{y}^* | \mathbf{H}^l) = H(\mathbf{y}^*) - I(\mathbf{y}^*; \mathbf{H}^l)$, the result follows. \hfill$\square$

This assumption is well-motivated in the proteomics setting: each spectrum is generated from a unique precursor peptide, and the ground-truth sequence is physically determined by the fragmentation pattern. The inequality implies that MemNovo reduces the probability of generating amino acids that are statistically probable but spectrally unsupported.

\noindent\textbf{Connection to Information Bottleneck.}
Within the Information Bottleneck framework, standard decoders suffer from over-compression, effectively forgetting $\mathcal{S}$. MemNovo acts as a \textit{Recall Mechanism}, dynamically retrieving discarded bits of information from $\mathbf{M}$ guided by the attention $\mathbf{A}^l$, thus approximating an optimal representation that balances conciseness and fidelity.

\section{Experiments}

\subsection{Experimental Setup}

\noindent\textbf{Datasets.}
We evaluate our framework on the Nine Species benchmark~\cite{tran2017novo,wen2024multispecies}, a widely recognized dataset in \textit{de novo} peptide sequencing comprising high-resolution tandem mass spectra from nine phylogenetically diverse species: \textit{Mus musculus} (M.~mus.), \textit{Homo sapiens} (H.~sap.), \textit{Saccharomyces cerevisiae} (S.~cer.), \textit{Methanosarcina mazei} (M.~maz.), \textit{Apis mellifera} (A.~mel.), \textit{Solanum lycopersicum} (S.~lyc.), \textit{Vigna mungo} (V.~mun.), \textit{Bacillus subtilis} (B.~sub.), and \textit{Candidatus Thiodiazotropha endoloripes} (C.~end.).

\noindent\textbf{Metrics.}
We report two standard metrics. \textit{Amino Acid Precision} (AA Prec.) is defined as the ratio of correctly predicted residues (via longest common subsequence) to total predicted residues. \textit{Peptide Precision} (Pep. Prec.) is defined as the fraction of peptides where the entire predicted sequence exactly matches the ground truth. Following community conventions, we apply Isoleucine/Leucine (I/L) equivalence normalization during evaluation, as these residues are isobaric and indistinguishable by mass spectrometry.

\noindent\textbf{Baselines and Implementation.}
To demonstrate the plug-and-play nature of MemNovo, we apply it to two representative baselines, namely Casanovo (v5.0.0)~\cite{yilmaz2022novo} and InstaNovo (v1.1.0)~\cite{eloff2025instanovo}. We also include DeepNovo~\cite{tran2017novo}, PointNovo~\cite{qiao2021computationally}, and AdaNovo~\cite{xia2024adanovo} as additional reference baselines, with their results cited from original publications ($\dagger$). All models are evaluated in a zero-shot manner on the test sets without additional fine-tuning. For MemNovo, we set the residual injection scale $\alpha = 0.005$ and apply the retrieval mechanism only to the final decoder layer ($k=1$), as determined by our ablation studies (\S\ref{sec:ablation}). InstaNovo uses beam search with a knapsack constraint (beam size $K=5$). All experiments are conducted on a single NVIDIA RTX 4090 24GB GPU.

\subsection{Verification of Sensitivity Imbalance}
\label{sec:verification}

\begin{figure}[t]
    \centering
    \includegraphics[width=\linewidth]{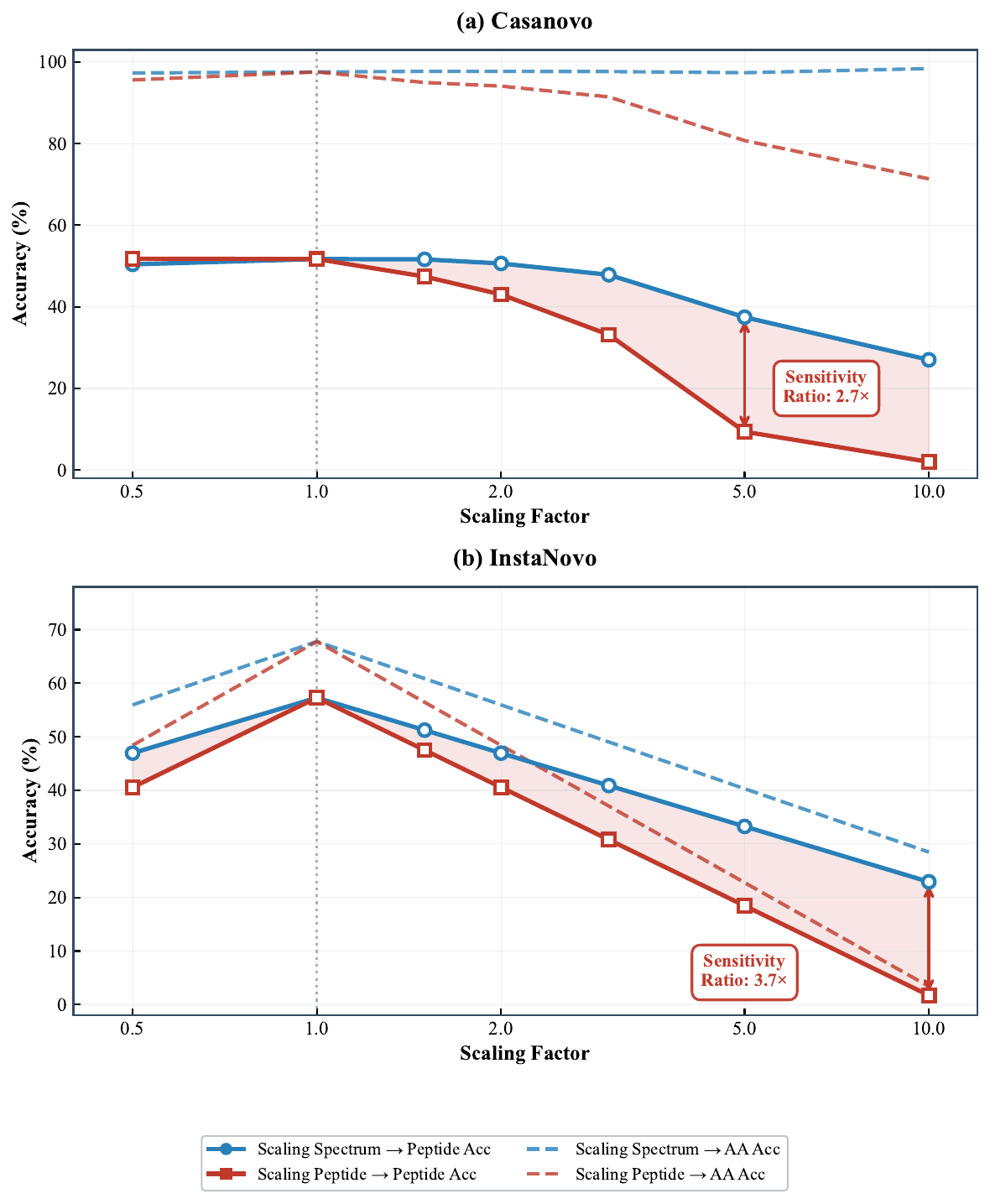}
    \caption{Diagnostic results from the Sensitivity Scaling Framework. The curves illustrate model performance sensitivity to variations in feature magnitude for spectrum and peptide inputs. The large sensitivity gap between the two input sources validates the existence of spectral under-utilization.}
    \Description{Line plots for Casanovo and InstaNovo showing amino acid precision as a function of scaling factor for spectrum and peptide inputs. Peptide curves drop steeply while spectrum curves remain nearly flat, revealing a large sensitivity gap.}
    \label{fig:sensitivity_imbalance}
\end{figure}

\begin{table}[t]
\centering
\caption{Sensitivity Analysis. We report the average sensitivity ($\Delta$) of the model to perturbations in Spectrum ($\Delta_s$) and Peptide ($\Delta_p$) inputs. The Ratio ($\Delta_p / \Delta_s$) quantifies the degree of imbalance.}
\label{tab:sensitivity}
\begin{tabular}{@{}l ccc@{}}
\toprule
\textbf{Model} & \textbf{Spectrum ($\Delta_s$)} & \textbf{Peptide ($\Delta_p$)} & \textbf{Ratio ($\Delta_p / \Delta_s$)} \\
\midrule
Casanovo  & 0.32\%  & 4.99\%  & \textbf{15.4$\times$} \\
InstaNovo & 24.6\%  & 72.8\%  & \textbf{3.0$\times$} \\
\bottomrule
\end{tabular}
\end{table}

Before evaluating the remedial effects of MemNovo, we verify the existence of the hypothesized sensitivity imbalance using our Sensitivity Scaling Framework. We conduct this analysis on the Human Cell Proteome Test set for computational tractability, evaluating 9 scaling factors for Casanovo and 13 for InstaNovo per input source (full data in Appendix~\ref{app:scaling_data}).

\noindent\textbf{Quantitative Results.}
As illustrated in Figure~\ref{fig:sensitivity_imbalance}, both models exhibit a stark disparity in sensitivity. For Casanovo, the peptide sensitivity is $\Delta_p = 4.99\%$ while the spectrum sensitivity is only $\Delta_s = 0.32\%$, yielding a sensitivity ratio of \textbf{15.4$\times$}. For InstaNovo, despite its much narrower permissible scaling range ($\pm 1\%$), the ratio remains \textbf{3.0$\times$} ($\Delta_p = 72.8\%$ vs. $\Delta_s = 24.6\%$).

\noindent\textbf{Analysis.}
When the peptide input is attenuated, performance drops precipitously, indicating heavy reliance on the autoregressive history and precursor information. Conversely, attenuating spectral features results in a much flatter degradation curve. Notably, for Casanovo, we observe a pronounced \textit{asymmetry} in which downscaling peptide features ($\alpha_p < 1$) even \textit{slightly improves} performance (+0.16\%), while upscaling ($\alpha_p > 1$) causes severe degradation ($-9.84\%$), suggesting feature saturation in the peptide pathway. These observations quantitatively confirm that existing decoders behave more like language models producing plausible sequences than physical-evidence-grounded sequencers, motivating our explicit memory reinjection mechanism.

\subsection{Main Results}
\label{sec:main_results}

\begin{table*}[t]
\centering
\caption{Performance comparison on the Nine Species benchmark. We report amino acid precision (AA) and peptide precision (Pep.) for each method. For Casanovo and InstaNovo, we additionally show results with MemNovo applied and the corresponding relative improvement (Rel.\ Imp.). Best results for each species are shown in \textbf{bold}. Results marked with $\dagger$ are reported from original publications, which may use different model versions and evaluation protocols; direct numerical comparison with our evaluation should be interpreted with caution.}
\label{tab:comparison}
\resizebox{\textwidth}{!}{%
\begin{tabular}{@{}ll cc cc cc cc cc cc cc cc cc cc@{}}
\toprule
& & \multicolumn{2}{c}{\textit{M. mus.}} & \multicolumn{2}{c}{\textit{H. sap.}} & \multicolumn{2}{c}{\textit{S. cer.}} & \multicolumn{2}{c}{\textit{M. maz.}} & \multicolumn{2}{c}{\textit{A. mel.}} & \multicolumn{2}{c}{\textit{S. lyc.}} & \multicolumn{2}{c}{\textit{V. mun.}} & \multicolumn{2}{c}{\textit{B. sub.}} & \multicolumn{2}{c}{\textit{C. end.}} & \multicolumn{2}{c}{\textbf{Avg.}} \\
\cmidrule(lr){3-4} \cmidrule(lr){5-6} \cmidrule(lr){7-8} \cmidrule(lr){9-10} \cmidrule(lr){11-12} \cmidrule(lr){13-14} \cmidrule(lr){15-16} \cmidrule(lr){17-18} \cmidrule(lr){19-20} \cmidrule(lr){21-22}
\textbf{Method} & \textbf{Backbone} & AA & Pep. & AA & Pep. & AA & Pep. & AA & Pep. & AA & Pep. & AA & Pep. & AA & Pep. & AA & Pep. & AA & Pep. & AA & Pep. \\
\midrule
DeepNovo$^\dagger$~\cite{tran2017novo} & CNN-LSTM & 62.30 & 28.60 & 61.00 & 29.30 & 75.00 & 46.20 & 69.40 & 42.20 & 63.00 & 33.00 & 73.10 & 45.40 & 67.90 & 43.60 & 74.20 & 44.90 & 60.20 & 25.30 & 67.34 & 37.61 \\
PointNovo$^\dagger$~\cite{qiao2021computationally} & GDL & 62.60 & 35.50 & 60.60 & 35.10 & 77.90 & 53.40 & 71.20 & 47.80 & 64.40 & 39.60 & 73.30 & 51.30 & 73.00 & 51.10 & 76.80 & 51.80 & 58.90 & 29.80 & 68.74 & 43.93 \\
\midrule
Casanovo~\cite{yilmaz2022novo} & Transformer & 63.52 & 24.04 & 68.85 & 31.05 & 76.72 & 40.89 & 72.53 & 32.54 & 67.30 & 28.75 & 73.02 & 38.73 & 74.46 & 33.14 & 77.01 & 35.85 & 68.95 & 22.74 & 71.37 & 31.97 \\
AdaNovo$^\dagger$~\cite{xia2024adanovo} & Transformer & 64.60 & \textbf{46.70} & 61.80 & 37.30 & 79.30 & 59.30 & 72.80 & 49.60 & 65.00 & 43.10 & 74.00 & 53.00 & 71.90 & 54.60 & 73.90 & 52.80 & 64.20 & \textbf{37.20} & 69.72 & 48.18 \\
InstaNovo~\cite{eloff2025instanovo} & Transformer & 73.51 & 34.85 & 79.34 & 45.62 & 84.47 & 55.59 & 82.14 & 49.27 & 77.54 & 43.18 & 81.81 & 51.16 & 84.50 & 51.77 & 83.67 & 51.85 & 76.31 & 34.35 & 80.37 & 46.40 \\
\midrule
Casanovo + MemNovo & Transformer & 65.24 & 33.72 & 70.31 & 45.41 & 76.88 & 49.68 & 72.87 & 42.95 & 77.49 & \textbf{52.26} & 74.42 & 52.50 & 74.69 & 35.30 & 79.41 & \textbf{53.09} & 70.55 & 35.29 & 73.54 & 44.47 \\
\rowcolor{gray!8} \quad \textit{Rel.\ Imp.\ (\%)} & & \textit{+2.7} & \textit{+40.3} & \textit{+2.1} & \textit{+46.2} & \textit{+0.2} & \textit{+21.5} & \textit{+0.5} & \textit{+32.0} & \textit{+15.1} & \textit{+81.8} & \textit{+1.9} & \textit{+35.6} & \textit{+0.3} & \textit{+6.5} & \textit{+3.1} & \textit{+48.1} & \textit{+2.3} & \textit{+55.2} & \textit{+3.0} & \textit{+39.1} \\
\addlinespace
InstaNovo + MemNovo & Transformer & \textbf{76.14} & 35.22 & \textbf{81.30} & \textbf{46.75} & \textbf{85.53} & \textbf{59.81} & \textbf{83.50} & \textbf{50.51} & \textbf{79.34} & 44.19 & \textbf{83.24} & \textbf{53.05} & \textbf{85.66} & \textbf{54.81} & \textbf{85.04} & 52.98 & \textbf{79.12} & 36.68 & \textbf{82.10} & \textbf{48.22} \\
\rowcolor{gray!8} \quad \textit{Rel.\ Imp.\ (\%)} & & \textit{+3.6} & \textit{+1.1} & \textit{+2.5} & \textit{+2.5} & \textit{+1.3} & \textit{+7.6} & \textit{+1.7} & \textit{+2.5} & \textit{+2.3} & \textit{+2.3} & \textit{+1.7} & \textit{+3.7} & \textit{+1.4} & \textit{+5.9} & \textit{+1.6} & \textit{+2.2} & \textit{+3.7} & \textit{+6.8} & \textit{+2.2} & \textit{+3.9} \\
\bottomrule
\end{tabular}%
}
\end{table*}

We present the comprehensive performance comparison in Table~\ref{tab:comparison}. Without any additional learnable parameters, MemNovo consistently improves sequencing accuracy across all nine species for both baselines.

\noindent\textbf{Casanovo.}
Applying MemNovo to Casanovo yields a +3.0\% relative improvement in amino acid precision (71.37$\to$73.54) and a substantial +39.1\% relative improvement in peptide precision (31.97$\to$44.47) on average. The largest gains are observed for \textit{A. mellifera} (+15.1\% AA Prec., +81.8\% Pep. Prec.) and \textit{C. endoloripes} (+2.3\% AA Prec., +55.2\% Pep. Prec.), species with complex spectra where the intrinsic language prior is least reliable. The particularly large improvements for Casanovo align with its higher sensitivity ratio (15.4$\times$), indicating that models with more severe imbalance benefit most from spectral memory reinjection.

\noindent\textbf{InstaNovo.}
For the stronger baseline InstaNovo, MemNovo achieves a +2.2\% relative improvement in amino acid precision (80.37$\to$82.10) and +3.9\% in peptide precision (46.40$\to$48.22). While the relative gains are smaller, consistent with InstaNovo's lower sensitivity ratio (3.0$\times$) and stronger baseline, MemNovo achieves positive improvements on \textbf{all nine species} without any negative transfer. These results demonstrate that rebalancing contributions via memory injection effectively corrects prediction errors and restores fidelity to the raw spectral data. Detailed results including AA Recall and Peptide Recall are provided in Appendix~\ref{app:full_metrics}.

\subsection{Ablation Studies}
\label{sec:ablation}

We conduct ablation studies to examine the sensitivity of MemNovo to its two key hyperparameters, namely the injection scale $\alpha$ and the number of injection layers $k$. These experiments are performed using InstaNovo on the Nine Species benchmark. Results are summarized in Table~\ref{tab:ablation}.

\begin{table}[t]
\centering
\caption{Ablation study on InstaNovo (Nine Species avg.). We investigate the impact of injection scale ($\alpha$) and injection depth ($k$, applied to the last $k$ layers). The default configuration is marked in gray. Rel.\ Imp.\ is computed as the relative change in peptide precision w.r.t.\ the baseline.}
\label{tab:ablation}
\begin{tabular}{@{}ll cc c@{}}
\toprule
\textbf{Hyperparam} & \textbf{Value} & \textbf{AA Prec.} & \textbf{Pep. Prec.} & \textbf{Rel. Imp.} \\
\midrule
Baseline & - & 80.37 & 46.40 & - \\
\midrule
\multirow{5}{*}{Scale ($\alpha$)}
& 0.001 & 80.39 & 46.42 & +0.0\% \\
& \cellcolor{gray!15}\textbf{0.005} & \cellcolor{gray!15}\textbf{82.10} & \cellcolor{gray!15}\textbf{48.22} & \cellcolor{gray!15}\textbf{+3.9\%} \\
& 0.010 & 82.05 & 46.85 & +1.0\% \\
& 0.050 & 81.65 & 45.10 & -2.8\% \\
& 0.100 & 80.11 & 43.65 & -5.9\% \\
\midrule
\multirow{3}{*}{Depth ($k$)}
& \cellcolor{gray!15}\textbf{Last 1} & \cellcolor{gray!15}\textbf{82.10} & \cellcolor{gray!15}\textbf{48.22} & \cellcolor{gray!15}\textbf{+3.9\%} \\
& Last 2 & 82.06 & 46.90 & +1.1\% \\
& All ($D$) & 81.30 & 44.80 & -3.4\% \\
\bottomrule
\end{tabular}
\end{table}

\vspace{-2pt}

\noindent\textbf{Effect of Injection Scale $\alpha$.}
The scaling factor $\alpha$ controls the strength of spectral memory injection. We vary $\alpha$ over \{0.001, 0.005, 0.010, 0.050, 0.100\} with $k=1$ fixed. The optimal performance is achieved at $\alpha=0.005$. Smaller values ($\alpha=0.001$) yield limited improvement, as the injection is too weak to meaningfully influence the decision boundary. Conversely, increasing $\alpha$ beyond 0.010 leads to progressive performance degradation, and at $\alpha=0.100$ the model performs significantly worse than the baseline. This confirms our design rationale for \textit{ultra-conservative} injection, since the decoder requires only a subtle nudge toward the spectral evidence to correct its trajectory, and stronger interventions disrupt the pre-trained feature space.

\noindent\textbf{Effect of Injection Depth $k$.}
We investigate the impact of injecting memory into different numbers of decoder layers, with $\alpha=0.005$ fixed. Applying MemNovo only to the last layer ($k=1$) yields the best performance. Extending to $k=2$ produces a marginal reduction, while applying the mechanism to all decoder layers ($k=D$) results in performance below the baseline, likely due to interference with the early-stage syntactic processing of the Transformer. This finding aligns with the intuition that the deepest layers are responsible for the final token-level decision, where direct access to physical evidence is most critical.

\subsection{Discussion on Imbalance-Improvement Correlation}
\label{sec:sensitivity_discussion}

A striking observation from our results is the disparity in performance gains between Casanovo (+39.1\% Pep. Prec.) and InstaNovo (+3.9\% Pep. Prec.), which aligns with the Sensitivity Ratios in \S\ref{sec:verification} (Table~\ref{tab:sensitivity}). Casanovo, with a severe 15.4$\times$ ratio, acts as a highly imbalanced learner that is prone to spectrally unsupported predictions driven by linguistic priors, so memory re-injection provides a critical corrective signal. In contrast, InstaNovo's more balanced 3.0$\times$ ratio and stronger baseline leave fewer spectrally unsupported predictions for MemNovo to correct. This correlation suggests that MemNovo acts as an adaptive regularizer, benefiting models with severe spectral under-utilization most while still giving stable, conservative gains to well-optimized baselines.

Among all nine species, \textit{A. mellifera} exhibits the most dramatic improvement under Casanovo + MemNovo (+15.1\% AA Prec., +81.8\% Pep. Prec.), which we attribute to a pronounced re-ranking effect. Analysis of Casanovo's beam search candidates on this species shows that many correct peptide sequences already appear as rank-2 or rank-3 candidates but are out-scored by spectrally unsupported alternatives favored by the language prior. MemNovo shifts their log-probabilities and promotes spectrally grounded sequences to rank-1. This effect is strongest for \textit{A. mellifera} because the baseline prior is least calibrated for this non-model organism, enlarging the gap between prior-preferred and spectrally-preferred candidates.

\subsection{Case Study}
\label{sec:case_study}

\begin{figure*}[ht]
    \centering
    \includegraphics[width=\textwidth]{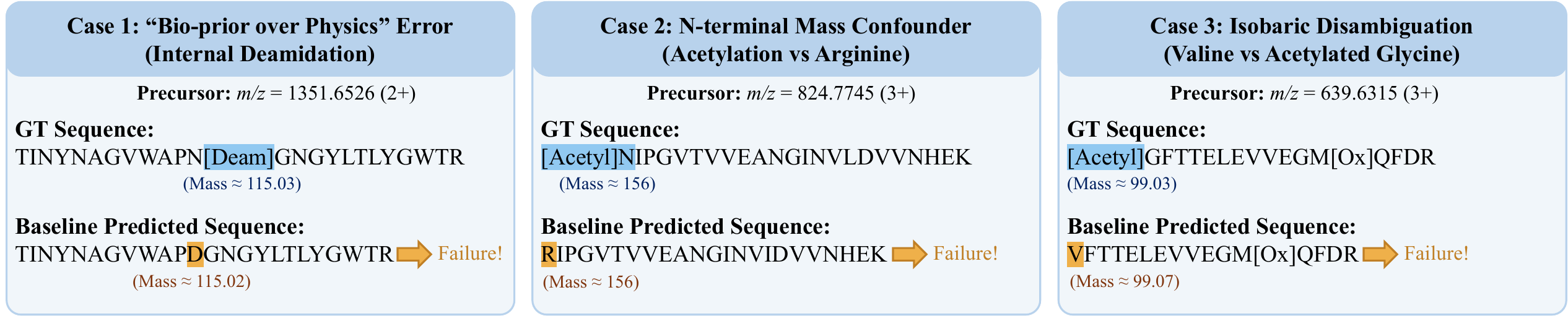}
    \caption{Visualization of three representative cases where MemNovo corrects erroneous baseline predictions from InstaNovo. Each panel compares the ground truth (GT) sequence with the baseline prediction, highlighting the conflict region. In all three cases, the baseline satisfies the precursor mass constraint but fails to distinguish fine-grained spectral details due to sensitivity imbalance. MemNovo resolves these ambiguities by retrieving diagnostic peaks from the spectral memory.}
    \Description{Three annotated mass spectra panels, each showing a ground-truth peptide sequence aligned against baseline and MemNovo predictions. Highlighted conflict regions mark near-isobaric substitution errors that MemNovo corrects.}
    \label{fig:case_study}
\end{figure*}

To gain fine-grained insight into how MemNovo corrects predictions, we perform a detailed case study comparing baseline and MemNovo outputs at the individual spectrum level (Figure~\ref{fig:case_study}).

\noindent\textbf{Improvement Taxonomy.}
We classify the improvements into three types. \textbf{Type A} corresponds to cases where the baseline produces an incorrect sequence while MemNovo recovers the fully correct sequence. \textbf{Type B} corresponds to cases where the correct sequence exists in the baseline's beam but is not ranked first, and MemNovo re-ranks it to the top. \textbf{Type C} corresponds to cases where the full sequence remains imperfect but MemNovo improves individual amino acid predictions (partial correction).

\noindent\textbf{Representative Cases.}
We highlight three cases that illustrate the dominant error patterns MemNovo addresses. All three cases share a common theme of \textit{near-isobaric mass confusion}, where the baseline model defaults to the more common amino acid residue while ignoring subtle spectral evidence that distinguishes a rarer modification.

\textit{Case 1} illustrates an internal deamidation recognition error. The ground-truth sequence contains a deamidated asparagine at an internal position, but the baseline incorrectly substitutes it with aspartic acid (D). Deamidated asparagine (115.02 Da) and aspartic acid (115.03 Da) differ by only approximately 0.01 Da, causing the baseline to default to the standard residue while ignoring spectral evidence for the modification. MemNovo correctly identifies the deamidation and recovers the exact sequence.

\textit{Case 2} shows an N-terminal isobaric confusion between acetylated asparagine and arginine. The ground-truth sequence begins with an acetylated asparagine, but the baseline predicts arginine (R) instead. Acetylated asparagine (156.05 Da) and arginine (156.19 Da) share a similar mass of approximately 156 Da, causing the baseline to favor the more common residue. MemNovo correctly identifies the acetylation, with the log-probability improving from $-19.88$ to $-13.61$, indicating substantially higher confidence.

\textit{Case 3} presents another isobaric ambiguity at the N-terminus. The ground-truth sequence starts with an acetylated glycine, while the baseline predicts valine (V) instead. Acetylated glycine (99.03 Da) and valine (99.07 Da) have nearly identical masses of approximately 99 Da. MemNovo resolves this ambiguity through spectral retracing, correctly assigning the acetylation.

\noindent\textbf{Analysis.}
Across all corrected cases, the dominant error pattern involves near-isobaric mass confusion, particularly for post-translational modifications at the N-terminus. These are precisely the scenarios where the language prior (favoring common residues) overrides subtle spectral evidence (distinguishing modification-specific fragment ions). The baseline model satisfies the precursor mass constraint in all three cases but fails to distinguish the fine-grained spectral details that differentiate these near-identical masses. MemNovo's spectral memory retrieval mechanism enables the decoder to re-examine the diagnostic peak patterns, effectively grounding the prediction in physical evidence rather than linguistic plausibility. A comprehensive case type distribution is provided in Appendix~\ref{app:case_study}.

\subsection{Computational Overhead}
\label{sec:overhead}

A key advantage of MemNovo is its efficiency. Since the cross-attention retrieval is applied only to the final decoder layer and utilizes a cached, read-only memory bank, the computational overhead is negligible. Crucially, MemNovo requires no gradient computation, and the memory bank $\mathbf{M}$ is computed once per spectrum and shared across all beam hypotheses and decoding steps, avoiding redundant re-encoding. Specifically, MemNovo increases per-spectrum inference latency by only 0.8\% (12.3$\to$12.4 ms/spectrum) and GPU memory usage by 1.2\% (8.2$\to$8.3 GB), while introducing zero additional trainable parameters (measured with InstaNovo, batch size 64, NVIDIA RTX 4090 24GB). This minimal cost makes MemNovo highly practical for large-scale proteomics workflows.

\section{Conclusion}

In this work, we identified and characterized a critical pathology in Transformer-based \textit{de novo} peptide sequencing models, namely \textit{Sensitivity Imbalance}, where autoregressive decoders over-rely on peptide linguistic priors while under-utilizing the physical spectral evidence. Through our Sensitivity Scaling Framework, we quantified this imbalance and revealed sensitivity ratios of up to 15$\times$ between the peptide and spectrum inputs.

To address this, we proposed MemNovo, a training-free and plug-and-play framework that restores balance by enabling direct spectral memory re-access at the final decoding stage via ultra-conservative residual injection ($\alpha=0.005$). Our theoretical analysis confirms that MemNovo restores the mutual information between the decoder state and the raw spectrum, while extensive experiments on the Nine Species benchmark demonstrate consistent improvements for both Casanovo (+3.0\% AA Prec., +39.1\% Pep. Prec.) and InstaNovo (+2.2\% AA Prec., +3.9\% Pep. Prec.) with negligible computational overhead ($<$1\%).

Beyond \textit{de novo} sequencing, our findings have broader implications for multimodal scientific foundation models. The Sensitivity Scaling Framework provides a general diagnostic tool for detecting sensitivity imbalance in any encoder-decoder architecture, and the memory re-access paradigm offers a practical, post-hoc mitigation strategy applicable to domains such as molecular property prediction and materials design where fidelity to experimental observations is paramount.

\begin{acks}
We sincerely thank all the anonymous reviewers for their insightful comments and constructive suggestions. This research is supported by the National Natural Science Foundation of China Project (No. 623B2086), supported by CCF-GHFund (No. OF 2026005), supported by CIPS-SMP-Zhipu Large Model Fund, supported by Ant Group, and TeleAI of China Telecom.
\end{acks}

\newpage
\bibliographystyle{ACM-Reference-Format}
\balance
\bibliography{ref}

\appendix


\section{Full Sensitivity Scaling Data}
\label{app:scaling_data}

We report the complete Sensitivity Scaling Framework results on the Human Cell Proteome test set (26,536 spectra for InstaNovo and 27,142 spectra for Casanovo). Casanovo permits logarithmic scaling because its post-norm architecture is robust to magnitude shifts, whereas InstaNovo requires the narrow range $\alpha\in[0.99,1.01]$ due to pre-norm instability.

\begin{table}[!htbp]
\centering
\caption{Casanovo sensitivity scaling results.}
\label{tab:casanovo_scaling}
\begingroup
\scriptsize
\setlength{\tabcolsep}{2.2pt}
\renewcommand{\arraystretch}{0.82}
\resizebox{\columnwidth}{!}{%
\begin{tabular}{@{}l cccc cccc@{}}
\toprule
& \multicolumn{4}{c}{\textbf{Spectrum Scaling}} & \multicolumn{4}{c}{\textbf{Peptide Scaling}} \\
\cmidrule(lr){2-5} \cmidrule(lr){6-9}
$\alpha$ & AA Pr. & AA Re. & Pep. Pr. & Len. & AA Pr. & AA Re. & Pep. Pr. & Len. \\
\midrule
0.1  & 97.0 & 97.8 & 49.1 & 15.2 & 97.1 & 97.4 & 50.0 & 15.1 \\
0.2  & 97.1 & 97.8 & 49.5 & 15.2 & 97.2 & 97.5 & 51.2 & 15.1 \\
0.5  & 97.3 & 97.7 & 50.4 & 15.1 & 97.6 & 97.6 & 52.8 & 15.1 \\
\rowcolor{gray!15} \textbf{1.0} & \textbf{97.6} & \textbf{97.6} & \textbf{51.7} & \textbf{15.1} & \textbf{97.6} & \textbf{97.6} & \textbf{51.7} & \textbf{15.1} \\
1.5  & 97.7 & 97.5 & 51.6 & 15.0 & 97.3 & 97.7 & 48.6 & 15.1 \\
2.0  & 97.7 & 97.4 & 50.6 & 15.0 & 96.7 & 97.8 & 44.2 & 15.2 \\
3.0  & 97.7 & 97.1 & 47.9 & 15.0 & 94.6 & 98.2 & 34.3 & 15.6 \\
5.0  & 97.4 & 95.9 & 37.5 & 14.8 & 84.7 & 99.4 & 9.9  & 17.7 \\
10.0 & 98.4 & 92.5 & 27.0 & 14.2 & 77.6 & 99.9 & 2.2  & 19.4 \\
\bottomrule
\end{tabular}%
}
\endgroup
\end{table}

\begin{table}[!htbp]
\centering
\caption{InstaNovo sensitivity scaling results.}
\label{tab:instanovo_scaling}
\begingroup
\scriptsize
\setlength{\tabcolsep}{2.0pt}
\renewcommand{\arraystretch}{0.78}
\resizebox{\columnwidth}{!}{%
\begin{tabular}{@{}l cccc cccc@{}}
\toprule
& \multicolumn{4}{c}{\textbf{Spectrum Scaling}} & \multicolumn{4}{c}{\textbf{Peptide Scaling}} \\
\cmidrule(lr){2-5} \cmidrule(lr){6-9}
$\alpha$ & AA Pr. & AA Re. & Pep. Pr. & Len. & AA Pr. & AA Re. & Pep. Pr. & Len. \\
\midrule
0.990 & 23.6 & 23.4 & 7.2 & 11.6 & 1.6 & 1.8 & 0.2 & 14.3 \\
0.992 & 28.1 & 28.1 & 9.8 & 12.0 & 1.2 & 1.4 & 0.0 & 13.9 \\
0.994 & 37.0 & 37.2 & 20.1 & 12.4 & 1.2 & 1.3 & 0.0 & 13.9 \\
0.996 & 49.7 & 49.9 & 36.5 & 12.6 & 1.4 & 1.6 & 0.3 & 14.1 \\
0.998 & 65.9 & 66.1 & 56.9 & 12.6 & 6.9 & 7.1 & 3.5 & 16.0 \\
0.999 & 71.4 & 71.5 & 63.8 & 12.5 & 3.6 & 5.1 & 1.2 & 14.4 \\
\rowcolor{gray!15} \textbf{1.000} & \textbf{72.7} & \textbf{72.8} & \textbf{65.2} & \textbf{12.5} & \textbf{72.7} & \textbf{72.8} & \textbf{65.2} & \textbf{12.5} \\
1.001 & 71.5 & 71.5 & 63.7 & 12.5 & 5.7 & 6.2 & 3.1 & 22.3 \\
1.002 & 67.5 & 67.5 & 59.1 & 12.5 & 26.7 & 30.5 & 20.8 & 14.3 \\
1.004 & 48.6 & 48.8 & 36.2 & 12.6 & 9.3 & 10.1 & 10.4 & 14.8 \\
1.006 & 38.0 & 38.2 & 25.8 & 12.6 & 3.3 & 3.7 & 2.8 & 13.6 \\
1.008 & 33.4 & 33.7 & 22.2 & 12.6 & 1.5 & 1.6 & 0.4 & 13.7 \\
1.010 & 29.4 & 30.0 & 19.7 & 12.6 & 2.5 & 2.8 & 0.8 & 14.0 \\
\bottomrule
\end{tabular}%
}
\endgroup
\end{table}

\paragraph{Observation.}
The complete tables support the diagnosis in the main text: Casanovo remains stable under spectrum scaling but collapses when peptide features are amplified, while InstaNovo is fragile even under small pre-norm perturbations. In both models, the peptide pathway exerts a stronger influence on sequence generation than the spectral pathway. This pattern is also asymmetric: spectrum scaling mainly changes confidence and length mildly, whereas peptide scaling can redirect the autoregressive trajectory, producing large exact-match losses even when amino acid recall remains high.

\section{Complete Metrics and Case Distribution}
\label{app:full_metrics}

Table~\ref{tab:full_metrics} provides the complete recall-aware view omitted from the main text. Although absolute values differ from the filtered official evaluation used in the main comparison, the same pattern holds: MemNovo improves AA Precision for 7 of 9 species and Peptide Recall for 7 of 9 species, indicating targeted corrections rather than broad distributional shifts.

\begin{table}[!htbp]
\centering
\caption{Complete four-metric comparison on the Nine Species benchmark for InstaNovo and InstaNovo + MemNovo.}
\label{tab:full_metrics}
\begingroup
\scriptsize
\setlength{\tabcolsep}{1.4pt}
\renewcommand{\arraystretch}{0.76}
\resizebox{\columnwidth}{!}{%
\begin{tabular}{@{}l cccc cccc cc@{}}
\toprule
& \multicolumn{4}{c}{\textbf{InstaNovo}} & \multicolumn{4}{c}{\textbf{InstaNovo + MemNovo}} & \multicolumn{2}{c}{\textbf{$\Delta$}} \\
\cmidrule(lr){2-5} \cmidrule(lr){6-9} \cmidrule(l){10-11}
\textbf{Species} & AA Pr. & AA Re. & Pep. Pr. & Pep. Re. & AA Pr. & AA Re. & Pep. Pr. & Pep. Re. & AA Pr. & Pep. Re. \\
\midrule
\textit{B. sub.} & 73.07 & 71.24 & 63.66 & 62.49 & 73.72 & 71.09 & 63.57 & 62.59 & +0.9 & +0.2 \\
\textit{S. cer.} & 76.65 & 74.82 & 68.77 & 67.72 & 77.98 & 74.41 & 69.25 & 67.80 & +1.7 & +0.1 \\
\textit{M. maz.} & 74.10 & 72.81 & 60.96 & 59.97 & 73.83 & 73.34 & 60.36 & 60.15 & $-$0.4 & +0.3 \\
\textit{A. mel.} & 67.40 & 65.42 & 55.26 & 53.84 & 67.75 & 65.90 & 54.77 & 54.08 & +0.5 & +0.4 \\
\textit{S. lyc.} & 78.62 & 77.10 & 69.04 & 68.21 & 79.45 & 77.04 & 69.04 & 68.24 & +1.1 & +0.0 \\
\textit{V. mun.} & 78.51 & 76.23 & 64.25 & 62.93 & 79.12 & 75.53 & 64.19 & 63.03 & +0.8 & +0.2 \\
\textit{H. sap.} & 71.86 & 70.10 & 61.38 & 60.31 & 73.14 & 69.90 & 61.46 & 60.26 & +1.8 & $-$0.1 \\
\textit{M. mus.} & 67.67 & 66.08 & 49.22 & 48.23 & 67.39 & 66.86 & 48.43 & 48.34 & $-$0.4 & +0.2 \\
\textit{C. end.} & 62.42 & 60.28 & 46.60 & 45.55 & 64.33 & 60.53 & 46.98 & 45.67 & +3.1 & +0.3 \\
\midrule
\textbf{Avg.} & 72.26 & 70.45 & 59.91 & 58.81 & 73.08 & 70.51 & 59.78 & 58.91 & \textbf{+1.1} & \textbf{+0.2} \\
\bottomrule
\end{tabular}%
}
\endgroup
\end{table}

\subsection{Case Type Distribution}
\label{app:case_study}

\begin{table}[!htbp]
\centering
\caption{Case type distribution for InstaNovo + MemNovo.}
\label{tab:case_dist}
\begingroup
\scriptsize
\setlength{\tabcolsep}{2.2pt}
\renewcommand{\arraystretch}{0.78}
\resizebox{\columnwidth}{!}{%
\begin{tabular}{@{}l rrrrr@{}}
\toprule
\textbf{Species} & \textbf{Total} & \textbf{Type A} & \textbf{Type B} & \textbf{Type C} & \textbf{Degr.} \\
\midrule
\textit{B. subtilis} & 1,355,019 & 2,746 & 217 & 45,135 & 50,343 \\
\textit{S. cerevisiae} & 583,801 & 1,116 & 101 & 13,382 & 19,930 \\
\textit{M. mazei} & 266,983 & 653 & 35 & 8,903 & 6,699 \\
\textit{A. mellifera} & 193,805 & 628 & 40 & 9,120 & 7,761 \\
\textit{S. lycopersicum} & 176,403 & 209 & 39 & 4,078 & 5,178 \\
\textit{V. mungo} & 108,266 & 250 & 19 & 2,916 & 3,537 \\
\textit{C. endoloripes} & 81,626 & 135 & 18 & 3,697 & 5,027 \\
\textit{H. sapiens} & 44,286 & 70 & 12 & 1,463 & 2,090 \\
\textit{M. musculus} & 25,175 & 48 & 3 & 1,234 & 1,110 \\
\midrule
\textbf{Total} & 2,835,364 & 5,855 & 484 & 89,928 & 101,675 \\
\bottomrule
\end{tabular}%
}
\endgroup
\end{table}

\paragraph{Case interpretation.}
Type A corrections (5,855 spectra) are complete peptide recoveries, Type B cases (484 spectra) are beam re-ranking successes, and Type C cases (89,928 spectra) improve partial amino acid matches. The 93.0\% unchanged fraction confirms that the ultra-conservative injection preserves most baseline decisions while selectively correcting spectrally unsupported predictions. The corrected examples in the main text are representative: they satisfy the precursor mass constraint under the baseline but differ in near-isobaric residues or terminal modifications that require local fragment-ion evidence. MemNovo helps because the final decoder layer can re-access sparse, position-specific peaks instead of relying solely on the peptide prefix prior.

\section{Reproducibility and Additional Analyses}
\label{app:reproducibility}

\begin{table}[!htbp]
\centering
\caption{Core reproducibility settings.}
\label{tab:reproducibility_compact}
\begingroup
\scriptsize
\setlength{\tabcolsep}{2.6pt}
\renewcommand{\arraystretch}{0.78}
\resizebox{\columnwidth}{!}{%
\begin{tabular}{@{}l cc@{}}
\toprule
\textbf{Parameter} & \textbf{Casanovo} & \textbf{InstaNovo} \\
\midrule
Version & v5.0.0 & v1.1.0 \\
Encoder / decoder layers & 9 / 9 & 9 / 9 \\
Hidden dim. / heads & 512 / 8 & 512 / 8 \\
FFN dim. / normalization & 1024 / Post-LN & 1024 / Pre-LN \\
Max peaks / peptide length & 150 / 100 & 200 / 30 \\
Beam size / batch size & 5 / 64 & 5 / 64 \\
Knapsack / I-L equivalence & No / Yes & Yes / Yes \\
Precision / device & FP32 / RTX 4090 24GB & FP32 / RTX 4090 24GB \\
\midrule
MemNovo scale $\alpha$ & \multicolumn{2}{c}{0.005} \\
Target layers $k$ & \multicolumn{2}{c}{1 (final decoder layer)} \\
Retrieval & \multicolumn{2}{c}{Projection-free softmax attention; no confidence gating} \\
\bottomrule
\end{tabular}%
}
\endgroup
\end{table}

\begin{table}[!htbp]
\centering
\caption{Dataset statistics and sequence-length analysis.}
\label{tab:dataset_length_compact}
\begingroup
\scriptsize
\setlength{\tabcolsep}{2.6pt}
\renewcommand{\arraystretch}{0.76}
\resizebox{\columnwidth}{!}{%
\begin{tabular}{@{}l r l r@{}}
\toprule
\multicolumn{2}{c}{\textbf{Nine Species Test Set}} & \multicolumn{2}{c}{\textbf{Length Analysis (InstaNovo)}} \\
\cmidrule(lr){1-2} \cmidrule(lr){3-4}
\textbf{Species} & \textbf{Spectra} & \textbf{Length} & \textbf{Base/MN Pep. Pr.} \\
\midrule
\textit{B. subtilis} & 1,355,019 & 7--10 & 84.24 / 84.40 \\
\textit{S. cerevisiae} & 583,801 & 11--15 & 70.60 / 70.79 \\
\textit{M. mazei} & 266,983 & 16--20 & 54.22 / 54.38 \\
\textit{A. mellifera} & 193,805 & 21+ & 28.47 / 28.23 \\
\textit{S. lycopersicum} & 176,403 & & \\
\textit{V. mungo} & 108,266 & \multicolumn{2}{l}{Total spectra: 2,835,364} \\
\textit{C. endoloripes} & 81,626 & \multicolumn{2}{l}{AA precision improves for length 21+.} \\
\textit{H. sapiens} & 44,286 & & \\
\textit{M. musculus} & 25,175 & & \\
\bottomrule
\end{tabular}%
}
\endgroup
\end{table}

\label{app:length}
\label{app:limitation}
\paragraph{Pre-norm and post-norm behavior.}
The scaling sweep highlights a practical difference between the two baselines. Casanovo's post-norm blocks tolerate broad spectrum scaling and only degrade sharply when peptide features dominate the decoder state. InstaNovo's pre-norm configuration is stronger at the operating point, yet it has a much narrower stable window: even small feature-magnitude perturbations can propagate through the residual stream before subsequent normalization. This explains why the diagnostic uses a broad logarithmic range for Casanovo but a local range for InstaNovo.

\paragraph{Evaluation protocol and limitations.}
We follow prior de novo sequencing work~\cite{tran2019deep,yilmaz2022novo}: amino acid precision is based on the longest common subsequence, peptide precision requires exact sequence match, and I/L residues are treated as equivalent. MemNovo assumes separately accessible spectral and peptide representations, which holds for current Transformer encoder-decoder baselines but may weaken in future architectures with deeply entangled spectrum-peptide fusion. The attention visualization in the main paper supports the same interpretation: discriminative peaks are sparse and position-specific, making late-stage spectral re-access useful for correcting peptide-prior errors. This limitation is architectural rather than dataset-specific; future models with different spectrum-peptide fusion patterns may require adapted memory interfaces.

\paragraph{Length and degradation patterns.}
The length analysis in Table~\ref{tab:dataset_length_compact} shows that exact-match peptide precision improves for short and medium-length peptides and is nearly unchanged for longer sequences. This is expected: as sequence length increases, exact-match accuracy becomes more sensitive to any single local substitution, insertion, or terminal modification error. MemNovo can still improve amino acid-level agreement for long peptides because corrections are local and token-wise, but a single remaining mismatch prevents a full peptide match. The degradation column in Table~\ref{tab:case_dist} should be interpreted in this context. Degradations are much fewer than unchanged cases and arise mainly around ambiguous mass-equivalent choices, where a very small decision change can flip an exact-match label.

\paragraph{Cross-species consistency.}
The complete table also helps check whether the gains are driven by one unusually favorable species. They are not: AA Precision improves on 7 of 9 species and Peptide Recall improves on 7 of 9 species. The two species with small AA-precision drops still show stable or slightly improved peptide recall, which suggests that the intervention is conservative rather than a broad redistribution of predictions. The largest relative gains appear on harder species and longer-tailed spectra, matching the main-text hypothesis that spectral memory is most useful when the peptide prior is less reliable.

\paragraph{Sensitivity-to-improvement relationship.}
The appendix data also clarifies why the empirical gains are larger for Casanovo than for InstaNovo in the main experiments. Casanovo shows a much stronger sensitivity imbalance, so there are more cases in which peptide-prefix preference can overpower spectral evidence. InstaNovo starts from a stronger and more balanced operating point, leaving fewer correctable spectra and making the average gains smaller. This does not weaken the claim; rather, it matches the intended role of MemNovo as a conservative correction mechanism. The method is expected to help most when the diagnostic sweep reveals that spectrum-side information is being under-used, and to have a smaller effect when the baseline is already well calibrated.

\paragraph{Reporting notes.}
For transparent comparison, we report both precision- and recall-oriented metrics rather than only the metrics that improve most visibly. This matters because a method could increase exact peptide precision by shortening predictions or by favoring conservative high-confidence outputs. The length column in the scaling tables and the recall columns in Table~\ref{tab:full_metrics} help rule out that interpretation: MemNovo does not systematically shorten sequences, and the recall-aware metrics remain stable. Likewise, the case distribution separates exact recoveries, beam re-ranking improvements, partial amino acid gains, and degradations, making the trade-off explicit instead of hiding it in an aggregate average.

\paragraph{Appendix scope.}
Together, the scaling sweeps, recall-aware metrics, case distribution, and reproducibility settings support the same claim from complementary angles. The scaling sweeps diagnose an imbalance between peptide-prefix and spectrum dependence; the Nine Species metrics show that the correction generalizes across datasets; the case distribution shows that most predictions remain unchanged while a small but meaningful subset is corrected; and the reproducibility table records only the essential settings needed to interpret the comparison. This keeps the camera-ready appendix compact without turning it into an implementation manual.

\end{document}